\documentclass[
]{ceurart}

\usepackage{listings}
\usepackage{xcolor} 
\usepackage{url}
\usepackage{hyperref}

\usepackage{graphicx}
\usepackage{subcaption} 
\usepackage[export]{adjustbox} 

\usepackage{pgfplotstable}
\usepackage{booktabs}

\usepackage{wrapfig}

\pgfplotsset{compat=1.18}

\begin{document}

\copyrightyear{2026}
\copyrightclause{Copyright for this paper by its authors.
  Use permitted under Creative Commons License Attribution 4.0
  International (CC BY 4.0).}

\conference{Workshop on GREEN-AI: Grounding Research on Energy- and resource-Efficient Networks for sustainable AI, Co-Located with CIKM 2026, November 08, 2026, Rome, Italy}

\title{Tug of Watts: A study on the Energy - Accuracy Trade-off of Large Language Models on Medical Question Answering Tasks}


\author[1,2]{Viju Sudhi}[%
email=viju.sudhi@uni-bielefeld.de,
]

\author[2,3]{Giulio Billi}[%
email=giulio.billi2@studio.unibo.it,
]
\author[2,3]{Andreea Scrob}[%
email=andreea.scrob@studio.unibo.it,
]
\author[1,2]{Regina Stodden}[%
email=regina.stodden@uni-bielefeld.de,
]
\author[1,2]{Philipp Cimiano}[%
email=cimiano@cit-ec.uni-bielefeld.de,
]
\address[1]{Cluster of Excellence Cognitive Interaction Technology (CITEC), Bielefeld, Germany}
\address[2]{Bielefeld University, Bielefeld, Germany}
\address[3]{University of Bologna, Bologna, Italy}





\begin{abstract}
Large Language Models (LLMs) are increasingly used in medical question answering (QA) systems. While recent works aim at incrementally improving the downstream task accuracy, an empirical estimation of the models' energy consumption and the associated carbon footprints is often overlooked. In this study, we demonstrate how practitioners can estimate energy consumption and arrive at reliable emission estimates for medical QA tasks available in the OpenMedical-LLM leaderboard by investigating 13 LLMs from 3 model families ranging from 270M to 14B parameters. We discuss how one can critically interpret energy versus accuracy trade-offs for a set of models for a given task. With this work, we underline that reporting disjoint energy and accuracy metrics alone does not suffice for careful trade-off assessments, but it also demands systematic construction and interpretation of joint trade-off metrics. We envisage such studies aiding practitioners in adopting \textit{better} and \textit{more efficient} models for their task and helping them become active participants in building sustainable LLM pipelines. 
\end{abstract}


\maketitle

\section{Introduction}

Large Language Models (LLMs) are increasingly used for a wide variety of tasks in different domains, including safety-critical ones like healthcare and medicine \cite{el-enen_med-llama3_2026, wang_large_2024, shool_systematic_2025, liu_survey_2026}. Recent studies show how frontier language models achieve passing scores on United States Medical Licensing Examination (USMLE) questions without specialized medical training and how they are able to provide clinically reasonable answers to a wide number of difficult medical scenarios \cite{nori_capabilities_2023, gilson_how_2023}. Attesting to this, there has been an increase in the number of domain-specific models, including Med-PaLM \cite{singhal_toward_2025}, MedGemma \cite{sellergren_medgemma_2026}, BioMistral \cite{labrak_biomistral_2024} and BioMedLM \cite{bolton_biomedlm_2024}, among others, which are further fine-tuned on medical and clinical data. These models report improved performance on different medical benchmarks than their corresponding general-purpose counterparts. 

While newer models are continuously (re-)trained to reach better performance on different downstream tasks, reporting energy costs and carbon footprints associated with them is often overlooked. 
Although LLM training costs are a one-time spend, the more important problem arises with the soaring inference demands from these models once they are deployed in real-time systems. Recent estimates suggest that this rising demand for AI inference would account for around 652 TWh of energy costs in 2030, compared to 23 TWh in 2022 \cite{noauthor_ai_2025}. 



The rising concerns with LLM energy demands have opened doors for ``efficiency-aware evaluation" of these models. Towards this goal, several tools have been developed to measure carbon footprints of LLMs, both at training and inference stages \cite{bannour_evaluating_2021}. These include CodeCarbon \cite{noauthor_mlco2codecarbon_2026}, and CarbonTracker \cite{anthony_carbontracker_2020}, among others. Despite researchers acknowledging that reporting energy costs adds more transparency to their pipelines, \textbf{we argue that merely reporting disjoint energy-accuracy metrics does not suffice}. 


We see this as a more convoluted problem both from an environmental and technological perspective. On one hand, while we acknowledge LLMs lead to escalated energy consumption, assessing models purely based on efficiency metrics is impractical. At the same time, the growing trend of improving models aimed only at better accuracy in benchmarks does not offer a convincing alternative either. In addition to this, recent studies \cite{maddox_rethinking_2020, muennighoff_scaling_2023, zadenoori_does_2025} show that model size does not always hold a linear correlation with model performance, which begs the question of why one should use a larger model with higher costs despite subpar performance. This motivates us to position our argument that \textbf{energy-accuracy measurements of LLMs should go hand in hand, complementing one another.} Practitioners should be aware of how to measure and interpret the trade-offs between these parameters, finally enabling them to make better and informed decisions about the choice of models for their task. In this context, we acknowledge that choosing the right model amidst the plethora of available models is itself \textit{``a sustainability decision"} \cite{noauthor_ai_2025} and to support this decision, we should treat the problem as a \textit{``multi-criteria decision making problem"} \cite{mehditabar_smart_2025} jointly optimizing the energy-accuracy metrics.










In this study\footnote{Our code is available here: \url{https://github.com/ag-sc/autollm}.}, using the example of Medical QA tasks in the OpenMedical-LLM leaderboard \cite{Medical-LLM-Leaderboard}, we demonstrate how practitioners can monitor the energy consumption during inference and arrive at reliable CO2 emission metrics using the package CodeCarbon \cite{noauthor_mlco2codecarbon_2026}. We then present how one can comprehend the energy versus accuracy trade-offs given a set of models on the given task. To this end, we study 13 models from 3 different model families (Qwen-3 \cite{yang_qwen3_2025}, Llama-3 \cite{grattafiori_llama_2024}, and Gemma-2/3 \cite{team_gemma_2024, team_gemma_2025}) within the range of 270M to 14B parameters. We conduct in-context learning with zero-shot settings and further fine-tune the models with LoRA parameter-efficient fine-tuning to study the effect of fine-tuning both on accuracy and energy consumption. We interpret the energy-accuracy trade-offs from these experiments with the help of a Pareto-optimal frontier, discuss the issues with such analyses, and extend our study to a more recent distance-based rating scheme, CIRC \cite{mehditabar_smart_2025}, to help interpret the trade-offs clearly and distinguish the models better.




\section{Related Work}

Over the recent years, efficiency-aware evaluation of models has gained particular traction. There are notable works measuring energy-accuracy trade-offs for different tasks and domains \cite{zschache_comparing_2025, tran_energy-aware_2025, islam_systematic_2025, samsi_words_2023}. \citet{maliakel_investigating_2025} investigate the energy-accuracy trade-offs in different tasks, including question answering, commonsense reasoning, and factual generation, while \citet{alizadeh_language_2025} conducted an extended study considering software development tasks, analyzing the trade-offs with Pareto frontiers. In contrast, \citet{mehditabar_smart_2025} argued that this is not the optimal approach, proposing more interpretable metrics to assess the trade-offs, offering better discriminatory levels.


A wide number of frontier models are trained on medical and clinical data to improve their performance on medical domain-specific tasks \cite{singhal_toward_2025, sellergren_medgemma_2026, labrak_biomistral_2024, bolton_biomedlm_2024}. Recent works explore parameter-efficient fine-tuning of models on medical data \cite{el-enen_med-llama3_2026} and compare them with full-weight fine-tuned models \cite{christophe_med42_2024}. Additionally, works investigate how quantized models can perform on medical domain tasks \cite{noauthor_how_nodate}. While the techniques presented in these works help reduce the carbon footprint of models, there has been very little systematic study in the medical domain to measure the energy-accuracy trade-offs and arrive at interpretable trade-off metrics. 


To the best of our knowledge, this is the first work that attempts to study the energy-accuracy trade-offs on medical QA tasks. Unlike other works, we extend our study to show how the trade-offs can be interpreted and critically analyzed for a more informed model selection procedure.

\section{Methodology}

\subsection{Energy and Emission Estimation}

\paragraph{Energy Estimation:}

For real-time monitoring of energy consumption, we resort to the widely used CodeCarbon package \cite{noauthor_mlco2codecarbon_2026}. CodeCarbon reads hardware power every 15s and integrates it over time to get the final energy consumption. This is computed as the product of the total power and the runtime of the process, given by: $Energy = \left(P_{GPU} + P_{CPU} + P_{RAM}\right) * Runtime$.

\vspace{5pt}
\noindent Here, $P_{GPU}$ is measured directly from the NVIDIA NVML reports. $P_{CPU}$ is measured by a power model driven by the CPU load, given as $TDP × (0.1 + 0.9 × load^3)$, where $load$ is the fraction of CPU capacity in use within the range of (0, 1) and $TDP$ is the processor’s rated power. We set $TDP$ to 450W, matching the CPUs in our infrastructure at 225W each. CodeCarbon estimates $P_{RAM}$ from the amount of installed memory and measures $Runtime$ as the time taken for inference.

\paragraph{Emission Estimation:}

CO2 emissions are computed as the product of energy and the carbon intensity of the local power grid. For example, in Germany, Carbon Intensity amounts to around 328 g  CO2eq/kWh according to the 2025 statistics \cite{noauthor_co2_nodate}. This is further scaled by a Power Usage Effectiveness factor (PUE) that accounts for data center overheads such as cooling. The total emission is given by: $\text{Emission} = Energy * \text{Carbon Intensity}  * PUE$.



\subsection{Energy versus Accuracy Trade-off}

To study the energy versus accuracy trade-offs, we consider two objectives: (i) minimize the energy consumption, and (ii) maximize the task performance. Hence, the ideal model should yield the maximum accuracy while consuming the least amount of energy.

\begin{figure}[htbp]
    \centering
    \includegraphics[width=0.35\linewidth, trim=10 10 10 10, clip]{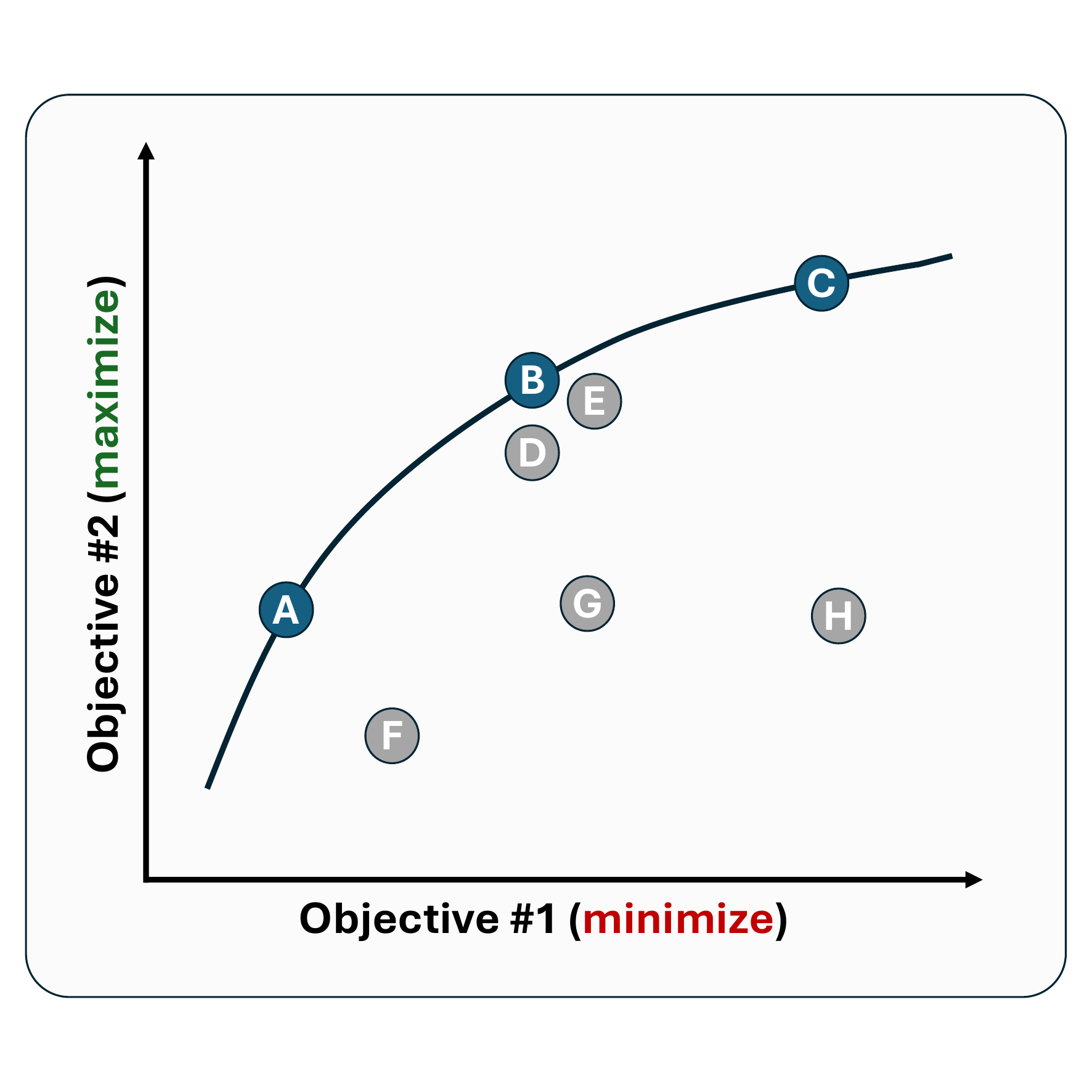}
    \caption{Example Pareto-optimal Frontier.}
    \label{fig:exampple_pareto_optimal_frontier}
\end{figure}

\paragraph{Pareto-optimal Frontier:} \citet{alizadeh_language_2025} analyze the relationship between energy consumption and accuracy of models with the help of Pareto fronts. These frontiers help in analyzing the balance between the defined objectives. However, Pareto fronts mark only the dominant points, leaving little to no interpretation for the rest of the non-dominant points. This hinders practitioners in distinguishing the contributions from the non-dominant points, limiting their options to the ones that lie on the Pareto front. For example, consider the exemplary Pareto front illustrated in Figure \ref{fig:exampple_pareto_optimal_frontier}. In this case, Objective \#1 should be minimized, and Objective \#2 should be maximized. The Pareto front shows the dominant points A, B, and C. While the points D and E are non-dominant, they are still close to B. The frontier gives no reasonable interpretation to distinguish between the non-dominant points, D and E, from one another and from the nearer dominant point, B. Consequently, one could only select from among the dominant points with the help of the Pareto front.


\paragraph{Joint metrics:} \citet{mehditabar_smart_2025} argue that a combination of both task accuracy metrics and energy consumption metrics is required for a more \textit{holistic} evaluation of real-world deployment constraints. Towards this, they introduced joint energy-accuracy metrics considering the problem as a Multi-Criteria Decision Making problem. One of their proposed rating schemes, ``Concentric Incremental Rating Circles" (CIRC), is based on Euclidean distance that quantifies each point by its distance to the most optimal point. This results in a clearer interpretation of all the points in the study and a clear distinction between one another. They use a five-point scale synonymous with the existing power ratings. We detail the computation of the CIRC ratings in Appendix \ref{appendix:circ-ratings}.


\section{Experiments}

\vspace{5pt}
\textbf{Dataset:} To study the energy versus accuracy trade-off of LLMs in medical QA tasks, we leveraged the publicly available Open Medical-LLM Leaderboard \cite{Medical-LLM-Leaderboard}. The leaderboard is based on different tasks including MedQA, MedMCQA, PubMedQA, and MMLU medical subsets. Together, these datasets provide a common evaluation setting for measuring medical knowledge and reasoning. 


\vspace{5pt}
\noindent \textbf{Models:} We ran the experiments on the following 13 models from 3 different model families. The models vary from 270M to 14B in parameter count.
\begin{itemize}
    \item Llama family \cite{grattafiori_llama_2024}: Llama-3.2-1B, Llama-3.2-3B, Llama-3.1-8B
    \item Gemma family \cite{team_gemma_2024, team_gemma_2025}: gemma-2-2b, gemma-2-9b, gemma-3-270m, gemma-3-1b-pt, gemma-3-4b-pt
    \item Qwen family \cite{yang_qwen3_2025}: qwen3-0.6b, qwen3-1.7b, qwen3-4b, qwen3-8b, qwen3-14b
\end{itemize}

\vspace{5pt}
\noindent \textbf{Pipeline:} We conducted our experiments as follows: (i) Firstly, we ran in-context learning experiments with zero-shot prompts on the pre-trained models; (ii) Further, we performed LoRA parameter-efficient fine-tuning on the models for a maximum of 5 epochs with early stopping. The evaluation step is then repeated on the fine-tuned models.


\vspace{5pt}
\noindent \textbf{Evaluation:} To evaluate the pre-trained models under a zero-shot setting and the LoRA fine-tuned models, we relied on the evaluation package \texttt{lm-eval-harness} \cite{eval-harness}. The package allows us to measure the model performance on the tasks and evaluate it against the gold references. In addition, we also estimated the energy consumption and CO2 emissions at the inference stage. 

\vspace{5pt}
\noindent \textbf{Hardware:} All training and evaluation runs are executed on a single NVIDIA A40 GPU with 48GB of memory. The node is equipped with two AMD EPYC 7713 CPUs.

\vspace{5pt}

\noindent More details on the datasets and evaluation metrics are provided in Appendix \ref{appendix:datasets}. Fine-tuning hyper-parameters are detailed in Appendix \ref{appendix:hyper-parameters}.

\begin{table}[t]
  \centering
  \caption{Accuracy, Energy Consumption and CO2 Emission Results}
  \begin{filecontents*}{data/energy_accuracy_results_latex_v1.csv}
model,pre_acc,sft_acc,pre_energy,sft_energy,pre_co2,sft_co2
qwen3-0.6b,48.73 $\pm$ 3.33,49.52 $\pm$ 3.33,0.022,0.031,0.008,0.012
qwen3-1.7b,54.83 $\pm$ 3.29,62.23 $\pm$ 3.18,0.023,0.036,0.009,0.014
qwen3-4b,68.56 $\pm$ 3.08,74.21 $\pm$ 2.75,0.04,0.054,0.015,0.02
qwen3-8b,67.13 $\pm$ 3.00,76.89 $\pm$ 2.65,0.066,0.064,0.025,0.024
qwen3-14b,\textbf{74.73 $\pm$ 2.74},\textbf{81.03 $\pm$ 2.45},0.077,0.089,0.029,0.034
gemma-2-2b,51.18 $\pm$ 3.30,56.21 $\pm$ 3.29,0.026,0.042,0.01,0.016
gemma-2-9b,69.06 $\pm$ 2.89,75.43 $\pm$ 2.69,0.06,0.082,0.023,0.031
gemma-3-270m,24.89 $\pm$ 2.85,27.33 $\pm$ 2.91,0.015,0.024,0.006,0.009
gemma-3-1b-pt,27.30 $\pm$ 2.79,30.17 $\pm$ 2.91,0.023,0.036,0.009,0.014
gemma-3-4b-pt,55.15 $\pm$ 3.27,63.98 $\pm$ 3.15,0.046,0.059,0.018,0.023
llama-3.2-1b,40.69 $\pm$ 3.28,43.80 $\pm$ 3.33,\textbf{0.012},\textbf{0.021},\textbf{0.005},\textbf{0.008}
llama-3.2-3b,60.87 $\pm$ 3.19,61.33 $\pm$ 3.22,0.022,0.038,0.009,0.014
llama-3.1-8b,69.59 $\pm$ 2.93,70.68 $\pm$ 2.89,0.038,0.06,0.015,0.023
\end{filecontents*}
\pgfplotstabletypeset[
    col sep=comma,
    string type, 
    columns/0/.style={column type/.try={l}},
    every head row/.style={
      before row={
        \toprule
        \multicolumn{1}{c}{\multirow{2}{*}{\textbf{Model}}} & \multicolumn{2}{c}{\textbf{Accuracy ($\uparrow$)}} & \multicolumn{2}{c}{\textbf{Energy (kWh) ($\downarrow$)}}  & \multicolumn{2}{c}{\textbf{CO2 Emission (kg) ($\downarrow$)}} \\
        \cmidrule(lr){2-3} \cmidrule(lr){4-5} \cmidrule(lr){6-7}
      },
      after row=\midrule
    },
    every last row/.style={after row=\bottomrule},
    display columns/0/.style={column name={}},  
    display columns/1/.style={column name={\textbf{Pre-trained}}},
    display columns/2/.style={column name={\textbf{Fine-tuned}}},
    display columns/3/.style={column name={\textbf{Pre-trained}}},
    display columns/4/.style={column name={\textbf{Fine-tuned}}},
    display columns/5/.style={column name={\textbf{Pre-trained}}},
    display columns/6/.style={column name={\textbf{Fine-tuned}}}    
  ]{data/energy_accuracy_results_latex_v1.csv}
  \label{tab:main-results}
\end{table}

\section{Results}

In Table \ref{tab:main-results}, we show the performance of all the models, pre-trained and fine-tuned, on the medical QA tasks. We also report the energy consumption and emissions per experiment. It is evident from the table that the largest model \texttt{qwen3-14b} yields the highest accuracy among all the models. This is true both for the pre-trained model and the fine-tuned counterpart. This model also results in the highest energy consumption and consequently the highest CO2 emission. The smallest model in the list, \texttt{gemma-3-270m}, yields the least accuracy on the task. Surprisingly, a slightly larger 1B parameter model, \texttt{llama-3.2-1b}, however, consumes the least energy.

\begin{figure}[htbp]
    \centering
    \includegraphics[width=\linewidth]{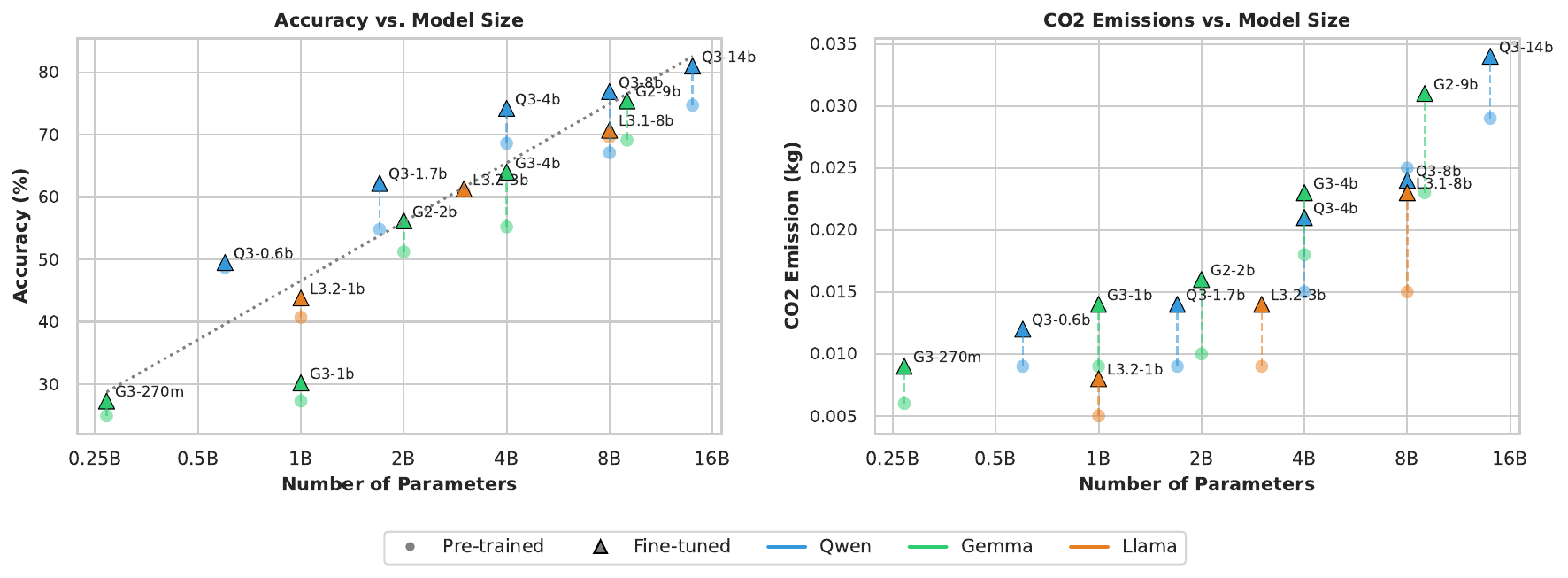}
    \caption{Accuracy and CO2 Emissions versus Model Size}
    \label{fig:accuracy-co2-emissions-versus-model-size}
\end{figure}

\paragraph{Effect of model size:}

As shown in Figure \ref{fig:accuracy-co2-emissions-versus-model-size}, for the medical QA tasks, there is a mostly linear relation between the accuracy and the model size. However, this is not the case when we compare emissions (energy) against the model size. Though smaller models generally emit less CO2 than the larger models, models belonging to the Llama family show a different trend. For example, a 1B parameter model \texttt{llama-3.2-1b} consumes less energy than a 600M parameter model \texttt{qwen-3-0.6b}. Similarly, an 8B parameter model \texttt{llama-3.1-8b} consumes the same amount of energy as a 4B parameter model \texttt{qwen3-4b}.

\paragraph{Low accuracy gains at the cost of high energy consumption:}

As illustrated in Table \ref{tab:delta-comparison}, we compute the relative performance gain and relative energy costs of the fine-tuned models per model family. For example, \texttt{qwen3-0.6b} yields an accuracy of 49.5 and \texttt{qwen3-14b} yields an accuracy of 81.0. This results in a relative accuracy gain of $(81.0-49.5) / 49.5 = +64\%$ for the Qwen3 family; however, at the cost of an 187\% increase in the energy cost. This is also the case for all the model families, where the improvement in task performance is only marginal when compared to the corresponding increase in energy consumption (and likewise, the increase in CO2 emission costs).

\begin{table}[htbp]
    \centering
    \caption{Comparison of relative accuracy gains versus relative energy costs}
    \begin{tabular}{lcc}
    \toprule
    \textbf{Model}  & \textbf{Relative Gain} & \textbf{Relative Costs}  \\
    \textbf{Family} & \textbf{(Accuracy)} & \textbf{(Energy)}  \\
    \midrule
    Qwen3 & +64\% & +187\% \\
    Gemma2 & +34\% & +95\% \\
    Gemma3 & +134\% & +146\% \\
    Llama3 & +61\% & +186\% \\
    \bottomrule
    \end{tabular}
    \label{tab:delta-comparison}
\end{table}

\begin{figure}
    \centering
    \includegraphics[width=\linewidth]{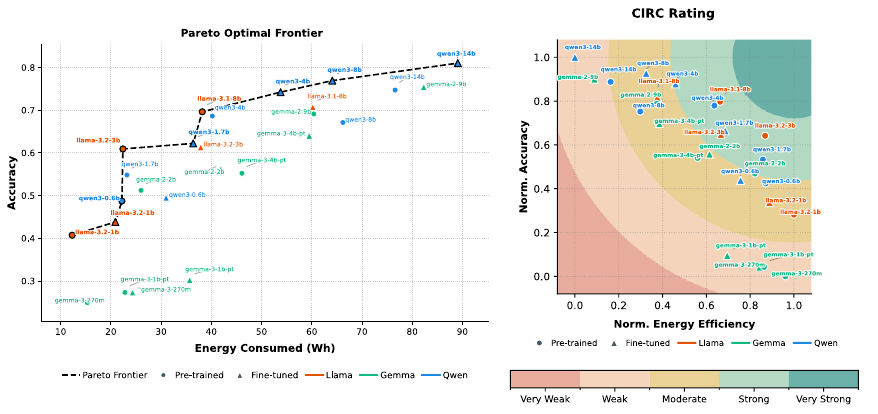}
    \caption{Interpreting Energy-Accuracy Trade-offs with Pareto-optimal frontier (left) and CIRC rating (right).}
    \label{fig:pareto_circ}
\end{figure}


\paragraph{Interpreting the Pareto-optimal front:} In Figure \ref{fig:pareto_circ} (left), we find that 9 models (out of 26) fall on the Pareto-optimal front. This includes 5 fine-tuned variants and 4 pre-trained ones. Neither the pre-trained nor the fine-tuned Gemma models are on the frontier. The plot shows diminishing returns in accuracy for the larger models, showing that the improvements in accuracy are only marginal but at the cost of relatively higher energy consumption. However, it is difficult to analyze how different \texttt{qwen3-1.7b} (PT) is compared with \texttt{llama-3.2-3b} (PT). This underlines the argument that although the Pareto frontier helps us to interpret the dominating points, it leaves no interpretation for the non-dominant points, consequently reducing the number of options to finally choose from.


\paragraph{Interpreting the CIRC ratings:} CIRC ratings illustrated in Figure \ref{fig:pareto_circ} (right) give a clearer interpretation of the energy-accuracy trade-offs. Here, energy efficiency is given by the complement of the measured energy consumption. The rating scheme assumes no model has reached the most optimal solution, leaving no model in the \textit{Very Strong} circle. In the \textit{Strong} circle, we find \texttt{llama-3.2-3b} (PT) and \texttt{llama-3.1-8b} (PT), the least distant from the optimal solution. 
Compared to the Pareto front, the extreme dominant points \texttt{llama-3.2-1b} (PT) and \texttt{qwen3-14b} (FT) belong to the \textit{Moderate} and the \textit{Weak} circles here, respectively. \texttt{qwen3-4b} (PT) is not a dominant point in the Pareto front, but is rated as \textit{Strong} just like the dominant point \texttt{llama-3.1-8b} (PT). 
This shows how one can make more informed choices with CIRC ratings offering a larger pool of choices, unlike Pareto-front marking only the dominant points.

\section{Discussion}

Our empirical evaluation of 13 models across three families demonstrates that relying solely on absolute accuracy or isolated efficiency metrics is fundamentally insufficient for selecting optimal LLMs in medical QA tasks. While larger models predictably achieve higher accuracy, their marginal performance gains often incur a disproportionate environmental cost. For instance, while \texttt{qwen3-14b (FT)} reaches a peak accuracy of 81.03\%, it requires 0.089 kWh and emits 0.034 kg of CO$_2$---an 187\% increase in energy overhead for only a 64\% relative accuracy gain over smaller baselines. 

Depending on operational constraints, practitioners face distinct deployment trade-offs: models like \texttt{llama-3.1-8b (PT)} offer balanced performance under moderate resource limits, whereas smaller fine-tuned models such as \texttt{qwen3-1.7b (FT)} serve as excellent high-throughput or edge alternatives. 

A key takeaway of our work is the limitation of traditional Pareto-optimal frontiers, which discard non-dominant models and offer no guidance for intermediate choices. By employing distance-based joint metrics like CIRC ratings, practitioners gain a more nuanced, multi-tier evaluation scale. This approach successfully identifies high-value, non-dominant options---such as \texttt{qwen3-4b (PT)} achieving a \emph{Strong} rating comparable to larger dominant architectures. 

Finally, while our primary evaluation focuses on inference-stage footprints, integrating fine-tuning energy into a life-cycle framework remains critical. Although parameter-efficient fine-tuning (LoRA) enhances medical QA performance, training overhead must be amortized over expected deployment lifetimes. Future work will explore hardware-level quantization and multimodal medical pipelines to further optimize sustainable AI deployments.




\section{Conclusion}

In this work, we demonstrated how to critically study the energy-accuracy trade-offs of 13 LLMs on medical QA tasks. We demonstrate how practitioners can measure the energy consumed in the inference process and relate this to the CO2 emissions per model per task. With our thorough evaluation, we show how larger models attain higher task performance, but at the cost of higher relative carbon footprints. We show how one can analyze the trends with the help of Pareto-optimal frontiers, discuss the problems with these settings, and study a distance-based rating scheme that helps to clearly distinguish the trade-offs of individual models.

\begin{acknowledgments}
This work was funded by the European Union and the German state of North Rhine-Westphalia within the following project of the European Regional Development Fund (EFRE-20801074): LLM4KMU - Optimized use of open-source large language models in SMEs.
\end{acknowledgments}

\section*{Limitations}

Energy estimates can vary based on the hardware specifications and inference optimizations. In this study, we resorted to a specific hardware resource for training and evaluation, with no further optimizations. In future work, gathering energy estimates from different hardware types at different optimization stages can help build comprehensive insights. For comparability in the final assessment of base models and fine-tuned models, we avoid the energy costs associated with fine-tuning. However, we note that these costs should be accommodated to help practitioners better distinguish the models suited for their task. This demands more granular reporting, and we defer this to future work. 

\section*{Declaration on Generative AI}

  
 
 During the preparation of this work, the authors used Grammarly to perform a grammar and spelling check. The authors also used Google Gemini for coding assistance to create the figures for the paper and minor rephrasing. After using these services, the authors reviewed and edited the content as needed and take full responsibility for the publication’s content. 

\bibliography{sample-ceur}

\appendix

\section{Datasets}
\label{appendix:datasets}

\paragraph{Fine-tuning.} For fine-tuning, we leveraged the train splits of all datasets in the OpenMedical-LLM leaderboard, except the MMLU subsets. To balance the number of samples per task, (i) we over-sample the training split of PubMedQA from 450 to 900 samples; (ii) we reduce the large MedMCQA dataset through stratified sub-sampling based on its 21 original medical domains, extracting a targeted subset of 20000 records. The details are presented in Table \ref{tab:datasets}.

\paragraph{Evaluation.} For evaluation, we use the test samples from all the datasets, including the MMLU subsets. This accounts for 3773 samples. For all datasets except MMLU subsets, we leverage the joint log-probability of the expected options. For MMLU subsets, the model response is evaluated based on exact match with the expected option.

\begin{table}[htbp]
    \centering
    \caption{Datasets included in Open Medical-LLM Leaderboard}
    \begin{tabular}{ll rrr l}
    \toprule
    \textbf{Dataset} & \textbf{Type} & \textbf{\# Train} & \textbf{\# Validation} & \textbf{\# Test} & \textbf{Evaluation} \\
    
    \textbf{} & \textbf{} & \textbf{Samples} & \textbf{Samples} & \textbf{Samples} & \textbf{Metric} \\    
    \midrule
    MedQA &  Open MCQA & 10178 & 1272 & 1273 & log-likelihood over\\
     &  &  &  &  & A / B / C / D\\
    
    MedMCQA &  Open MCQA & 18000 & 4183 & 2000 & log-likelihood over\\
     &  &  &  &  & A / B / C / D\\
     
    PubMedQA &  Closed MCQA & 900 & 50 & 500 & log-likelihood over\\
     &  &  &  &  & yes / no / maybe\\
     
    MMLU Subsets & Open MCQA & - & - & 1089 & exact match\\
     &  &  &  &  & on the option word\\
     
    \midrule
    & \textbf{Total} & 29078 & 5505 & 4862 \\
    \bottomrule
    \end{tabular}
    
    \label{tab:datasets}
\end{table}

\section{Hyper-parameters}
\label{appendix:hyper-parameters}

For LoRA fine-tuning of the models, we keep the base model weights in \texttt{bfloat16} and update only the additional low-rank adapters. We compute the loss only on the completion tokens, masking the input tokens. The effective batch size is kept at 32 across all models. All runs use a learning rate of $10^{-4}$ with 10 warm-up steps. Training is allowed to run for up to 5 epochs, but the model is evaluated periodically on the validation split during training, and the validation loss is monitored through early stopping. Training stops when the validation loss does not improve by at least $10^{-4}$ for 5 consecutive evaluations. This serves two purposes: it limits overfitting on the
relatively small supervised datasets, and it avoids spending training energy
on epochs that no longer improve generalization.

\section{Energy consumption and CO2 emissions during fine-tuning}
\label{appendix:fine-tuning-estimates}

Table \ref{tab:fine-tuning-estimates} details the energy consumption and carbon emissions associated with fine-tuning the models. Resource requirements scale with model size and architecture; the lightweight \texttt{gemma-3-270m} incurs the lowest footprint with a runtime of 754.59 seconds, consuming 0.069 kWh and emitting 0.026 kg of CO2. Conversely, the larger \texttt{qwen3-14b} model demands the highest resources, requiring 27,956.03 seconds, 2.752 kWh of energy, and resulting in 1.048 kg of CO2 emissions.

\begin{table}[htbp]
    \centering
    \caption{Estimates of energy consumption and CO2 emission for fine-tuning runs}    
    \begin{tabular}{lcccc}
    \toprule
    \multirow{2}{*}{\textbf{Model}} & \textbf{Runtime} & \textbf{Energy}  & \textbf{CO2 Emission} \\
    \textbf{} & \textbf{(s)} & \textbf{(kWh)}  & \textbf{(kg)} \\

    \midrule
    qwen3-0.6b & 2767.698 & 0.255 & 0.097 \\
    qwen3-1.7b & 3704.089 & 0.353 & 0.134 \\
    qwen3-4b & 10920.611 & 1.071 & 0.408 \\
    qwen3-8b & 20619.091 & 1.986 & 0.756 \\
    qwen3-14b & 27956.034 & 2.752 & 1.048 \\
    gemma-2-2b & 6490.905 & 0.617 & 0.235 \\
    gemma-2-9b & 14251.333 & 1.367 & 0.521 \\
    gemma-3-270m & 754.591 & 0.069 & 0.026 \\
    gemma-3-1b-pt & 2862.367 & 0.257 & 0.098 \\
    gemma-3-4b-pt & 8505.431 & 0.820 & 0.312 \\
    llama-3.2-1b & 2140.442 & 0.205 & 0.078 \\
    llama-3.2-3b & 4559.037 & 0.450 & 0.172 \\
    llama-3.1-8b & 10842.395 & 1.072 & 0.408 \\
    \bottomrule
    \end{tabular}

    \label{tab:fine-tuning-estimates}
\end{table}

\section{Extended Evaluation Results}

\begin{table}
\caption{Evaluation results of pre-trained models}
\label{tab:pre_trained_results}
\begin{tabular}{lccccc}
\toprule
\textbf{Model} & \textbf{PubMedQA} & \textbf{MedMCQA} & \textbf{MedQA} & \textbf{MMLU Subsets} & \textbf{Avg. Accuracy} \\
\midrule
qwen3-0.6b & 70.20 $\pm$ 2.05 & 38.82 $\pm$ 0.75 & 38.57 $\pm$ 1.36 & 48.50 $\pm$ 3.88 & 48.73 $\pm$ 3.33 \\
qwen3-1.7b & 72.60 $\pm$ 2.00 & 48.10 $\pm$ 0.77 & 48.08 $\pm$ 1.40 & 54.11 $\pm$ 3.83 & 54.83 $\pm$ 3.29 \\
qwen3-4b & 75.80 $\pm$ 1.92 & 57.09 $\pm$ 0.77 & 62.06 $\pm$ 1.36 & 70.34 $\pm$ 3.56 & 68.56 $\pm$ 3.08 \\
qwen3-8b & 76.20 $\pm$ 1.91 & 62.47 $\pm$ 0.75 & 65.83 $\pm$ 1.33 & 66.62 $\pm$ 3.48 & 67.13 $\pm$ 3.00 \\
qwen3-14b & 76.40 $\pm$ 1.90 & \textbf{66.34 $\pm$ 0.73} & \textbf{71.25 $\pm$ 1.27} & \textbf{76.43 $\pm$ 3.18} & \textbf{74.73 $\pm$ 2.74} \\
gemma-2-2b & 74.00 $\pm$ 1.96 & 40.86 $\pm$ 0.76 & 35.04 $\pm$ 1.34 & 51.78 $\pm$ 3.85 & 51.18 $\pm$ 3.30 \\
gemma-2-9b & \textbf{78.60 $\pm$ 1.84} & 57.92 $\pm$ 0.76 & 60.02 $\pm$ 1.37 & 70.84 $\pm$ 3.37 & 69.06 $\pm$ 2.89 \\
gemma-3-270m & 37.60 $\pm$ 2.17 & 25.05 $\pm$ 0.67 & 22.86 $\pm$ 1.18 & 23.08 $\pm$ 3.28 & 24.89 $\pm$ 2.85 \\
gemma-3-1b-pt & 64.20 $\pm$ 2.15 & 31.05 $\pm$ 0.72 & 24.98 $\pm$ 1.21 & 20.92 $\pm$ 3.18 & 27.30 $\pm$ 2.79 \\
gemma-3-4b-pt & 72.60 $\pm$ 2.00 & 46.31 $\pm$ 0.77 & 45.64 $\pm$ 1.40 & 55.30 $\pm$ 3.81 & 55.15 $\pm$ 3.27 \\
llama-3.2-1b & 57.80 $\pm$ 2.21 & 37.03 $\pm$ 0.75 & 35.35 $\pm$ 1.34 & 39.34 $\pm$ 3.80 & 40.69 $\pm$ 3.28 \\
llama-3.2-3b & 73.00 $\pm$ 1.99 & 49.44 $\pm$ 0.77 & 51.77 $\pm$ 1.40 & 62.27 $\pm$ 3.72 & 60.87 $\pm$ 3.19 \\
llama-3.1-8b & 75.60 $\pm$ 1.92 & 56.28 $\pm$ 0.77 & 59.94 $\pm$ 1.37 & 72.41 $\pm$ 3.41 & 69.59 $\pm$ 2.93 \\
\bottomrule
\end{tabular}
\end{table}

\begin{table}
\caption{Evaluation results of fine-tuned models}
\label{tab:sft_results}
\begin{tabular}{lccccc}
\toprule
\textbf{Model} & \textbf{PubMedQA} & \textbf{MedMCQA} & \textbf{MedQA} & \textbf{MMLU Subsets} & \textbf{Avg. Accuracy} \\
\midrule
qwen3-0.6b & 59.40 $\pm$ 2.20 & 43.08 $\pm$ 0.77 & 38.96 $\pm$ 1.37 & 50.71 $\pm$ 3.86 & 49.52 $\pm$ 3.33 \\
qwen3-1.7b & 72.40 $\pm$ 2.00 & 50.01 $\pm$ 0.77 & 48.78 $\pm$ 1.40 & 64.82 $\pm$ 3.70 & 62.23 $\pm$ 3.18 \\
qwen3-4b & 77.80 $\pm$ 1.86 & 60.29 $\pm$ 0.76 & 63.08 $\pm$ 1.35 & 77.79 $\pm$ 3.16 & 74.21 $\pm$ 2.75 \\
qwen3-8b & 79.40 $\pm$ 1.81 & 64.36 $\pm$ 0.74 & 68.50 $\pm$ 1.30 & 79.96 $\pm$ 3.06 & 76.89 $\pm$ 2.65 \\
qwen3-14b & 79.80 $\pm$ 1.80 & \textbf{68.04 $\pm$ 0.72} & \textbf{72.27 $\pm$ 1.26} & \textbf{84.86 $\pm$ 2.78} & \textbf{81.03 $\pm$ 2.45} \\
gemma-2-2b & 75.60 $\pm$ 1.92 & 44.23 $\pm$ 0.77 & 41.08 $\pm$ 1.38 & 57.50 $\pm$ 3.85 & 56.21 $\pm$ 3.29 \\
gemma-2-9b & \textbf{80.40 $\pm$ 1.78} & 59.67 $\pm$ 0.76 & 63.94 $\pm$ 1.35 & 79.14 $\pm$ 3.11 & 75.43 $\pm$ 2.69 \\
gemma-3-270m & 50.00 $\pm$ 2.24 & 26.94 $\pm$ 0.69 & 24.82 $\pm$ 1.21 & 24.04 $\pm$ 3.34 & 27.33 $\pm$ 2.91 \\
gemma-3-1b-pt & 70.60 $\pm$ 2.04 & 32.58 $\pm$ 0.72 & 27.18 $\pm$ 1.25 & 23.52 $\pm$ 3.33 & 30.17 $\pm$ 2.91 \\
gemma-3-4b-pt & 77.20 $\pm$ 1.88 & 49.10 $\pm$ 0.77 & 52.95 $\pm$ 1.40 & 66.10 $\pm$ 3.68 & 63.98 $\pm$ 3.15 \\
llama-3.2-1b & 62.60 $\pm$ 2.17 & 34.76 $\pm$ 0.74 & 33.07 $\pm$ 1.32 & 43.97 $\pm$ 3.87 & 43.80 $\pm$ 3.33 \\
llama-3.2-3b & 76.80 $\pm$ 1.89 & 47.98 $\pm$ 0.77 & 50.12 $\pm$ 1.40 & 62.84 $\pm$ 3.75 & 61.33 $\pm$ 3.22 \\
llama-3.1-8b & 78.80 $\pm$ 1.83 & 56.49 $\pm$ 0.77 & 60.33 $\pm$ 1.37 & 73.41 $\pm$ 3.37 & 70.68 $\pm$ 2.89 \\
\bottomrule
\end{tabular}
\end{table}

Tables \ref{tab:pre_trained_results} and \ref{tab:sft_results} present the extended model evaluation results across the different datasets in the Open Medical-LLM Leaderboard on the pre-trained and fine-tuned models, respectively. Overall accuracy is averaged across the individual datasets, and the corresponding standard error is reported as the root-mean-square (RMS). In most of the cases, fine-tuning yields performance improvements across models and datasets. Larger models experience the most pronounced gains; for example, \texttt{qwen3-14b} model increases its average accuracy from 74.73\% to 81.03\%. Dataset-specific analysis shows that while models maintain strong baseline performance on PubMedQA, more challenging datasets like MedMCQA, MedQA, and MMLU Subsets benefit substantially from fine-tuning, with \texttt{qwen3-14b} achieving peak scores of 68.04\%, 72.27\%, and 84.86\%, respectively.

\section{Extended CIRC Ratings}
\label{appendix:circ-ratings}

For a given task, CIRC ranks the models based on their Euclidean distance to the optimal solution. To this end, firstly,  we normalize the accuracy ($Acc$) and energy consumption ($Energy$) values. Further, we calculate ``energy efficiency" as the complement of ``energy consumed" as: $Eff_{norm} = 1 - Energy_{norm}$. This helps us employ similar trends (higher the better) in both energy-accuracy axes. Finally, we compute the distance as $\sqrt{(1-Acc_{norm})^2 + (1-Eff_{norm})^2}$ for each model from the optimal solution.

To visualize this, we construct five concentric circles, each representing a CIRC rating, centered at the point $(1, 1)$. The point  represents the most optimal solution with maximum accuracy and maximum energy efficiency. These concentric circles have radii increasing in equal increments, starting from $\frac{\sqrt{2}}{5}$ (representing a rating of 5) and ending at $\sqrt{2}$ (representing a rating of 1). For more details of the algorithm, we defer to the original work \cite{mehditabar_smart_2025}. Table \ref{tab:circ-ratings} presents the CIRC ratings for the 13 pre-trained and 13 fine-tuned models in our experiments. 

\begin{table}[htbp]
  \centering
  \caption{CIRC Ratings}
  \begin{filecontents*}{data/circ_rating.csv}
model,Training,Accuracy,Norm. Accuray,Energy (kWh),Norm. Energy,Distance,Rank
qwen3-0.6b,Pre-trained,48.7,0.424,0.022,0.871,0.590,3
qwen3-0.6b,Fine-tuned,49.5,0.439,0.031,0.757,0.612,3
qwen3-1.7b,Pre-trained,54.8,0.533,0.023,0.858,0.488,4
qwen3-1.7b,Fine-tuned,62.2,0.665,0.036,0.686,0.459,4
qwen3-4b,Pre-trained,68.6,0.779,0.040,0.637,0.425,4
qwen3-4b,Fine-tuned,74.2,0.879,0.054,0.460,0.554,4
qwen3-8b,Pre-trained,67.1,0.752,0.066,0.298,0.744,3
qwen3-8b,Fine-tuned,76.9,0.927,0.064,0.326,0.678,3
qwen3-14b,Pre-trained,74.7,0.888,0.077,0.163,0.845,3
qwen3-14b,Fine-tuned,81.0,1.000,0.089,0.000,1.000,2
gemma-2-2b,Pre-trained,51.2,0.469,0.026,0.822,0.560,4
gemma-2-2b,Fine-tuned,56.2,0.558,0.042,0.615,0.586,3
gemma-2-9b,Pre-trained,69.1,0.788,0.060,0.374,0.661,3
gemma-2-9b,Fine-tuned,75.4,0.900,0.082,0.089,0.917,2
gemma-3-270m,Pre-trained,24.9,0.000,0.015,0.961,1.001,2
gemma-3-270m,Fine-tuned,27.3,0.043,0.024,0.844,0.970,2
gemma-3-1b-pt,Pre-trained,27.3,0.043,0.023,0.863,0.967,2
gemma-3-1b-pt,Fine-tuned,30.2,0.094,0.036,0.695,0.955,2
gemma-3-4b-pt,Pre-trained,55.2,0.540,0.046,0.560,0.637,3
gemma-3-4b-pt,Fine-tuned,64.0,0.697,0.060,0.385,0.685,3
llama-3.2-1b,Pre-trained,40.7,0.282,0.012,1.000,0.718,3
llama-3.2-1b,Fine-tuned,43.8,0.337,0.021,0.888,0.672,3
llama-3.2-3b,Pre-trained,60.9,0.642,0.022,0.868,0.382,4
llama-3.2-3b,Fine-tuned,61.3,0.649,0.038,0.667,0.484,4
llama-3.1-8b,Pre-trained,69.6,0.797,0.038,0.663,0.394,4
llama-3.1-8b,Fine-tuned,70.7,0.816,0.060,0.376,0.650,3
\end{filecontents*}
\pgfplotstabletypeset[
    col sep=comma,
    string type, 
    every head row/.style={
      before row={
        \toprule
      },
      after row=\midrule
    },
    every last row/.style={after row=\bottomrule},
    display columns/0/.style={column name={\textbf{Model}}},
    display columns/1/.style={column name={\textbf{Mode}}},
    display columns/2/.style={column name={\textbf{$Acc$}}},
    display columns/3/.style={column name={\textbf{$Acc_{norm}$}}},
    display columns/4/.style={column name={\textbf{$Energy$}}},
    display columns/5/.style={column name={\textbf{$Eff_{norm}$}}},
    display columns/6/.style={column name={\textbf{Distance}}}, 
    display columns/7/.style={column name={\textbf{Rating}}} 
  ]{data/circ_rating.csv}
  \label{tab:circ-ratings}
\end{table}

\paragraph{Sensitivity Analysis.} CIRC relies on min-max normalization to map accuracy ($Acc$) and energy efficiency ($Eff$) to a $[0, 1]$ scale and applies equal weighting in the distance computation. To evaluate the sensitivity of our findings to these methodological choices, we analyze variations along two dimensions:
\begin{itemize}
    \item \textbf{Normalization:} While min-max normalization preserves relative distances, it can be sensitive to outliers (e.g., extreme models like \texttt{gemma-3-270m} or \texttt{qwen3-14b}). Substituting min-max scaling with rank-based percentile normalization or z-score scaling yields stable top-tier rankings. Specifically, models identified in the \textit{Strong} rating tier (such as \texttt{llama-3.2-3b (PT)} and \texttt{llama-3.1-8b (PT)}) remain within the top performance cluster, as their trade-off profiles consistently balance high accuracy without excessive energy consumption.
    \item \textbf{Weighting:} Equal weighting assumes accuracy and energy efficiency carry equal importance. In real-world medical deployments, practitioner priorities can skew. If energy efficiency is weighted more heavily (e.g., weight of $0.7$ on energy and $0.3$ on accuracy), smaller efficient models like \texttt{qwen3-1.7b (FT)} shift inward to higher rating circles. Conversely, prioritizing clinical accuracy (e.g., $0.7$ weight on accuracy) naturally pulls high-performing large models like \texttt{qwen3-8b (FT)} closer to the optimal center. However, the intermediate tiers remain largely stable, confirming that CIRC provides a resilient multi-criteria categorization without being overly brittle to moderate weight adjustments.
\end{itemize}


\end{document}